\documentclass[letterpaper]{article}
\usepackage{aaai24}
\usepackage{times}
\usepackage{helvet}
\usepackage{courier}

\usepackage[hyphens]{url} 
\usepackage{graphicx} 
\usepackage{graphicx} 
\usepackage{natbib} 
\usepackage{caption} 
\usepackage{booktabs}

\begin{document}
%
\title{Digits micro-model for accurate and secure transactions}

\author {
    Chirag Chhablani\textsuperscript{\rm 1,\thanks{Work was done as part of PhD summer internship at Vail Systems, Inc.}},
    Nikhita Sharma\textsuperscript{\rm 2,\thanks{Work was done while author was employed by Vail Systems, Inc.}},
    Jordan Hosier\textsuperscript{\rm 2},
    and Vijay K. Gurbani\textsuperscript{\rm 2,\rm 3}
}
\affiliations{
    \textsuperscript{\rm 1}Department of Computer Science, University of Illinois, Chicago\\
    \textsuperscript{\rm 2}Vail Systems, Inc.\\
    \textsuperscript{\rm 3}Department of Computer Science, Illinois Institute of Technology\\
    cchhab2@uic.edu, nikhitasharma2606@gmail.com, jhosier@vailsys.com, vgurbani@\{vailsys.com, iit.edu\}
}


\maketitle
\begin{abstract}

Automatic Speech Recognition (ASR) systems are used in the financial domain to enhance the caller experience by enabling natural language understanding and facilitating efficient and intuitive interactions.  The increasing use of ASR systems requires that such systems exhibit very low error rates.  The predominant ASR models to collect numeric data are large, general-purpose models provided commercially --- Google Speech-to-text (STT), or Amazon Transcribe --- or available through open source (OpenAI's Whisper).  Such ASR models are trained on hundreds of thousands of hours of audio data and require considerable resources to run.  Despite recent progress in such large speech recognition models, in this work we highlight the potential of smaller, specialized \textit{``micro''} models. Such light models can be trained perform well on number recognition specific tasks, competing with general models like Whisper or Google STT while using less than 80 minutes of training time and occupying at least an order of less memory resources. Also, unlike larger speech recognition models, micro-models are trained on carefully selected and curated datasets, which makes them highly accurate, agile, and easy to retrain, while using low compute resources. We present our work on creating micro models for multi-digit number recognition that handles diverse speaking styles reflecting real-world pronunciation patterns.  Our work contributes to domain-specific ASR models, improving digit recognition accuracy, and maintaining privacy of data.  An added advantage of our micro-models is their low resource consumption allows them to be hosted on-premise, thereby keeping private data local instead of sending it to an external cloud for processing.  Our results indicate that our micro-model makes less errors than the best-of-breed commercial or open-source ASRs in recognizing digits (1.8\% error rate of our best micro-model versus 5.8\% error rate of Whisper), and has a low memory footprint (0.66 GB VRAM for our model versus 11 GB VRAM for Whisper).


\end{abstract}

\section{Introduction}
\begin{figure}[h]
  \includegraphics[width=.45\textwidth]{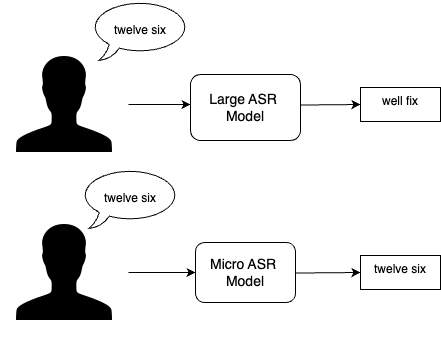}
  \caption{Figure shows the effect of large and micro language model on ambiguous inputs. Due to large vocabulary the large language model can lead to erroneous outputs whereas the micro models can generate an output which is within the range of provided vocabulary}
  \label{fig:asr-audio}
\end{figure}

Today, many financial transactions take place over the phone. In this setting, deploying accurate speech recognition systems is crucial for precise digit recognition. The accuracy with which machine learning models recognize multi-digit utterances plays a pivotal role in shaping the efficiency of various financial applications. For example, a scenario in which a voice-enabled transaction involves a series of digits that are mistranscribed due to the limitations of a domain-general model, may result in financial discrepancies or authentication failures. Figure \ref{fig:asr-audio}, shows an example of text where the utterance "twelve six" is transcribed incorrectly as "well fix" because of large domain of the large ASR model whereas it is transcribed correctly because of small domain of micro ASR model. The consequences could range from minor errors to significant financial discrepancies and bad user experience. Moreover in this context of voice-enabled financial transactions, the importance of deploying models that go beyond mere efficiency to ensure precise digit recognition becomes paramount. 

The significance of accurate digit recognition extends beyond efficiency to include privacy considerations - particularly in financial domains that deal with credit numbers, CVV or account numbers. Accurate STT APIs like Google-speech APIs require sending the data to an external cloud that can have significant privacy concerns.  This is where the imperative for socially responsible, privacy-aware micro-models comes into play. Consider a micro-model specifically tailored for financial contexts and adept at recognizing spoken digits with a heightened emphasis on privacy. Such a model not only ensures the accurate processing of financial transactions but also prioritizes the protection of personal financial information by avoiding an external API call. Further by addressing pronunciation variations and potential misinterpretations, this micro-model exemplifies the fusion of accuracy and efficiency. In this work, we focus on training a multi-digit micro-model that can handle diverse production of digits with high accuracy and computational efficiency. 

There is a noticeable research gap in the open-source datasets available for training models on spoken digit recognition. Most prior datasets either focus solely on single-digit utterances or predominantly on multi-digit utterances where digits are spoken individually. Our work bridges this gap by including a wide range of sequences representing various digit lengths and variations in pronunciation. For example, ``653'' may be spoken as ``six hundred fifty-three,'' ``six fifty-three,'' or ``sixty-five three.'' Specifically, the curated dataset is comprised of 14,000 utterances and aims to address the lack of diversity in the production of spoken digit data. This dataset is publicly available\footnote{The dataset can be downloaded from\\ https://github.com/chiraguic/SpokenMultiDigitVarietyDataset} and can be used by the research community to train speech recognition models to recognize credit card and account numbers - cases in which the data is not generally available due to its highly sensitive nature. 

Morever, the utilization of external APIs may raise privacy concerns due to the potential training of the ASR model on user data\footnote{A representative example of a privacy statement is https://cloud.google.com/speech-to-text/docs/data-logging} . While external providers often assert non-ownership of private data, the process of training their models relies on large datasets, including user inputs. This indirect exposure of sensitive information to external entities poses a risk to user privacy; published literature documents successful identification of individuals from data used to train models \cite{inbook,yin2023defeating,10.1145/2810103.2813677,10.1145/3448016.3457234}. In contrast, the adoption of micro-models in an in-house capacity offers a more controlled and socially-responsible environment. With micro-models, organizations maintain ownership and oversight of the training process, reducing the likelihood of inadvertent data exposure. This approach not only enhances privacy but also provides greater transparency and control over how user data is handled, mitigating the potential risks associated with external API usage.

Our contributions can be summarized as follows: 
\begin{enumerate}
    
    \item We highlight the potential of micro-models over domain-general models for specialized use cases.
    \item We create --- and make publicly available at the URL shown in the footnote --- a dataset for multi-digit recognition that can handle diverse articulation of digits for upto five-digit numbers.
    \item We train two micro-models that outperform domain-general commercial  ASR models like Google and Whisper \cite{radford2023robust} in terms of recognition accuracy, and require a fraction of resources to train and deploy them.
\end{enumerate}

The rest of the paper is structured as follows: We briefly summarize how ASR systems work in the next section.  Related work and experiment details follow to outline the curation of our dataset and the techniques used for training the models.  We discuss the results next followed by a conclusion.

\section{ASR Background}

The objective of an Automatic Speech Recognition (ASR) system is to convert an audio signal into its corresponding text, known as a  \emph{transcript}.  The transcript is subsequently used by back-end systems to drive a specific application.  As an example, consider an account holder calls into a financial institution, and the system challenges her by asking an account number. The ASR captures the acoustic signal corresponding to user's utterance as she speaks out the digits and converts, or decodes, it to a textual transcript consisting of the digits spoken during authentication.  The transcript is then presented to a back end system that retrieves the account based on the digits contained in the transcript.

ASR systems typically include an acoustic model, a language model, and a decoder. This system processes speech input to produce the most accurate transcription of spoken words. In short, the feature extraction module isolates pertinent features from the speech signal, minimizing unnecessary noise. The acoustic model captures speech acoustics and transcribes the extracted audio features into a sequence of context-dependent phonemes, often using deep neural networks in contemporary models. The language model determines the likelihood of specific words or word sequences based on surrounding context. Finally, the decoder utilizes the acoustic model, grammar, and language model in tandem to generate probable word sequences for a given audio frame, with the highest probability word sequence being the final text output. A comprehensive exploration of deep neural network-based ASR systems is presented in the work of Roger et al. \cite{roger2022deep}.

To delve further into the process, the input waveform undergoes segmentation into small frames, typically 25 milliseconds in duration, and from these frames, specific features are extracted. Commonly utilized features include Mel-Frequency Cepstral Coefficients (MFCCs), Cepstral Mean and Variance Normalization (CMVNs) representing audio content, or i-Vectors capturing speaker or utterance style. The selected features must effectively capture human speech characteristics while minimizing unwanted noise. This compression of the audio signal results in a sequence of fixed-length vectors through feature extraction. Subsequently, the acoustic model predicts the phoneme spoken in each audio frame. The acoustic model is tasked with modeling speech acoustics and transcribing the extracted audio features into a sequence of context-dependent phonemes. Training acoustic models involves Deep Neural Networks (DNNs) processing extensive datasets, typically comprising thousands of hours of human-transcribed audio data.

Finally, in (ASR) systems, a lexicon or dictionary links each word to its phonetic representation. This mapping helps convert predicted phonemes into words and, ultimately, full sentences. The language model plays a crucial role in determining the likelihood of specific words or word sequences based on surrounding context. This context is typically generated by neural networks or n-gram models trained on extensive textual datasets. The decoder then utilizes the acoustic model, grammar, and language model together to generate word sequences for a given audio frame. The final text output is the word sequence with the highest probability.

\section{Related Work}
Several open-source datasets exist in the realm of speech processing for digit recognition, each having its own set of strengths and limitations. The \textit{MNIST Speech Dataset} curates a dataset by embedding spoken digit recognition into the well-known MNIST format \cite{mnist_speech}. However, a notable constraint of this dataset is that it is focused solely on single-digit numbers. It does not address the challenges posed by multi-digit sequences containing a great deal of variation in production style. For instance, numbers like ``480'' can be produced as ``four hundred eighty'' or ``four hundred and eighty,'' variations that are not accounted for in the dataset.

In contrast, the \textit{Snips SLU Dataset}, \textit{Fluent Speech Commands dataset}, and \textit{SLURP dataset} focus on speech command recognition \cite{saade2019spoken,lugosch2019speech,bastianelli2020slurp}. While these datasets contribute to real-world applications, they don't encompass the nuances of recognizing numbers spoken in varied forms or numbers embedded within sentences. These limitations are shared by the \textit{Google Speech Commands Dataset}, which excels in recognizing short spoken commands but falls short in adequately handling multi-digit variations and numbers spoken within larger contexts.

Furthermore, even when datasets cover multi-digit utterances, they lack variety. For example, the \textit{Timers-and-such Dataset} offers an intriguing avenue for exploring temporal event classification and spoken timer detection \cite{lugosch2021timers}. However, it exhibits limitations, such as fewer utterances for numbers from ten to nineteen (10-19), which are crucial for recognizing the complete spectrum of five-digit numbers. Additionally \textit{Aurora Dataset} focuses on standalone multi-digit utterances \cite{auroradataset}. However, a key limitation of the Aurora dataset is that it records the digits spoken one at a time, potentially not addressing the variations in pronunciation seen in real-world scenarios.

Domain-specific datasets, like those detailed above, as well as domain-specific ASR models have garnered significant attention in recent research. 
For instance, \cite{jha2021learning}, targets domain-specific models for cases where domain-specific data is available in different languages. \cite{bekal2021remember} focuses on the accurate recognition of domain-specific words and named entities (i.e. addresses, names, etc.) More recently, \cite{de2023political} present experiments using a domain-specific language model for political speeches. \cite{dong2023speech} propose a complex but effective speech recognition method based on a domain-specific language speech network (DSL-Net) and a confidence decision network (CD-Net). They test their results on various open-source medical domain datasets.  

Unlike the prior work outlined in this section, we focus on creating a domain-specific micro-model for commonly occurring digit recognition tasks that can be used in any situation, for instance, the financial domain. While existing datasets and models trained on them contribute significantly to digit recognition endeavors, there remains a gap in handling standalone multi-digit utterances spoken in diverse ways. Addressing this gap is particularly useful to enhance caller experiences in financial transactions where accurate and flexible multi-digit recognition is vital. Our proposed dataset and the associated micro-model trained on it aim to tackle this challenge, providing improved accuracy and adaptability to variable pronunciations.

\section{Experimental Details}
\subsection{Dataset}
We curated the proposed dataset by harnessing data available from the Timers and Such \cite{lugosch2021timers} and LibriSpeech datasets \cite{lugosch2019speech}. The Timers and Such dataset is an open-source dataset of spoken English commands for common voice control use cases involving numbers. It has four intents, corresponding to four common offline voice assistant uses: SetTimer, SetAlarm, SimpleMath, and UnitConversion.  The dataset is fairly small, with 2,151 non-synthetic utterances, but it is considered useful for experimentation. The LibriSpeech dataset is a corpus of approximately 1000 hours of read English speech with a sampling rate of 16 kHz. It is derived from read audiobooks from the LibriVox project and has been carefully segmented and aligned . Our objective was to meticulously extract spoken numbers from sentences present in these datasets, ultimately constructing a comprehensive repository for flexible multi-digit recognition. Extending prior work on single and double-digit numbers, we focus on numbers up to five digits in length. We compiled an exhaustive vocabulary comprising all tokens necessary for articulating five-digit numbers, encompassing numerical digits and number-related phrases. These tokens include single digits from zero to nine, double digits tokens from ten to nineteen, and tokens like ``twenty'', ``thirty,'' etc. as shown in Table \ref{tab:words}.  In addition to digits, tokens representing key numerical expressions like "hundred" and "thousand" constitute the foundational building blocks for constructing complex numeric sequences. Further, we include the tokens ``and'', ``Oh'', and ``O'' as they are often used interchangeably with the digit zero in conversational speech. 

Subsequently, we employed state-of-the-art Whisper models \cite{radford2023robust} to glean precise timestamps of these tokens from the audio samples. The Whisper model delivered a detailed JSON file containing words and their corresponding timestamps (for e.g. "words": [
{"text": "eighty","start": 0.5, "end":0.9,"confidence": 0.90
}])\footnote{It was observed that JSON outputs often contain a slight offset in the start and end times for words, sometimes capturing a very small portion of the previous or next word. Nevertheless, it appears that this has minimal impact on the performance of the model.}.
We first identified the timestamps corresponding to the words of interest (for e.g. "four"-0:38-0:40s, "five"-0:40-0:42, etc). We then use the timestamp information for the each word to combine continuous sequences of digits (for e.g. "four five"-0:38-0:42)  to assemble all multi-digit numbers articulated within the sentence. Finally, after extracting all relevant words from the audio, we append one second of silence at the beginning and end of the audio files. This step ensures that the individual audio files are of sufficient length for training.

\begin{table}[h]
\centering
\begin{tabular}{c|c|c}
\multicolumn{3}{c}{\textbf{List of Tokens in the Vocabulary}} \\
\hline
One & Eleven & Twenty \\
Two & Twelve & Thirty \\
Three & Thirteen & Forty \\
Four & Fourteen & Fifty \\
Five & Fifteen & Sixty \\
Six & Sixteen & Seventy \\
Seven & Seventeen & Eighty \\
Eight & Eighteen & Ninety \\
Nine & Nineteen & Hundred \\
 Oh & O & 
Ten \\
Thousand & And & Zero \\ 
\end{tabular}
\caption{Vocabulary}
\label{tab:words}
\end{table}

The Kaldi toolkit \cite{povey2011kaldi} was used for training the ASR model.  Kaldi is a free, open-source toolkit for speech recognition research.  It is considered best-of-breed open source ASR training toolkit and has been used as the foundation for many commercial ASR systems.

\subsubsection{Train and Test Datasets}
To train the model we also included the Aurora dataset \cite{auroradataset}  to ensure that the model can recognize sequential as well as non-sequential digits. Refer to the appendix about brief introduction on Aurora 5 dataset. We adopted a train-test ratio of 90:10 over the entire dataset.  The dataset boasts a cumulative duration of about 4 hours. To standardize the audio data and ensure compatibility with phone audio recordings, all files were converted to an 8 kHz sample rate and encoded in 16-bit PCM format. The contribution from each dataset is shown in Table \ref{data:datainfo}. The number of audio files and corresponding hours of data contributed by each of the datasets is specified.
\begin{table}[h]
\centering
\begin{tabular}{lccc}
\hline
Dataset & Train files (hrs) & Test files (hrs) \\
\hline
Aurora-5 & 7421 (2.30) & 1355 (0.28) \\
Timers-and-such & 2502 (0.61) & 211 (0.10) \\
Librispeech & 2721 (0.72) & 230 (0.12) \\
\hline
\end{tabular}
\caption{Dataset Information}
\label{data:datainfo}
\end{table}

\subsection{Micro-model Architecture}
We trained two neural network based digit models- \textit{Micro-model-dense} and \textit{Micro-model-light} with varied architectures. The dense model is more accurate but demonstrates slightly higher latency during decoding.   

The input of the \textit{Micro-model-dense} network comprises two components: an i-vector input with a dimensionality of 100 and a raw input with a dimensionality
of 40. The network commences its transformation by applying an affine operation using an LDA matrix to the concatenated spliced frames of the i-vector, thereby reducing its dimensionality while retaining essential information. The subsequent stages involve a series of carefully orchestrated layers that progressively extract and manipulate features.
A fully connected layer, characterized by ReLU activation, batch normalization, and dropout, introduces non-linearity and regularization of the data. Following this, a sequence of 13 time-delay neural network (TDNN) layers are introduced, leveraging factorized structures for computational efficiency and capturing temporal dependencies. These TDNN layers, each comprising 1,536 dimensions with a bottleneck dimension of 160 and varying time strides, enable the network to capture intricate temporal patterns in the input data. Continuing the transformation, a linear layer processes the features, and a pre-final layer undertakes non-linear transformations with dimension reduction, thereby molding the data to match the requirements of specific modeling tasks. Here, the architecture bifurcates into two distinct branches. One branch caters to chain modeling, involving another pre-final layer and an output layer. The pre-final layer reduces the dimensions from 1,536 to 256, while the output layer generates predictions without the application of log-softmax transformation. This configuration is tailored for tasks necessitating sequence labeling. Simultaneously, the second
branch addresses cross-entropy training, encompassing analogous pre-final and output layers. These layers facilitate learning through cross-entropy loss, and the output layer produces classifications.

The \textbf{Micro-model-dense} neural network architecture, while achieving a lower Word Error Rate (WER\footnote{The WER is a widely accepted standard measure of ASR performance; it is expressed as a value between [0, 1.0] or as a percentage.  Lower values are better.  Please see the Appendix for more information on the WER.}), necessitates a slightly higher decoding time and demands more space. In an effort to strike a better balance between accuracy, space efficiency, and decoding time, we opted for a simplification of the neural network, dubbing it \textbf{Micro-model-light}. A significant adjustment involved a meticulous reduction in the number of TDNNF layers from 13 to 4, effectively mitigating continuous and resource-intensive weight multiplication operations. Additionally, recognizing the potential impact of excessively high dimensions on the space and time intensity of the model, we deliberately chose to judiciously decrease the dimensionality of the network's internal representations from 1,536 to 1,024. This architectural streamlining resulted in reduction in decoding time, which is more noticeable for CPU based decoding. For GPU based decoding, the difference is marginal. The simpler design leads to slight increase in the WER.

\begin{table}[ht]
\centering
\begin{tabular}{lcc}
\toprule
                  & \textbf{\begin{tabular}[c]{@{}c@{}}Micro-model-\\ light\end{tabular}} & \textbf{\begin{tabular}[c]{@{}c@{}}Micro-model-\\ dense\end{tabular}} \\ 
\midrule
WER                  & 2.6\%                                                                & 1.8\%                                                                \\
Model size (GPU, VRAM)  & 0.64 GB                                                          & 0.66 GB                                                                 \\
Model size (CPU, RAM)  & 0.16 GB                                                          & 0.20 GB                                                                 \\
Latency (RTF) on CPU & 0.00343                                                                 & 0.01432                                                                 \\
Latency (RTF) on GPU & 0.00061                                                               & 0.00060                                                               \\ 
\bottomrule
\end{tabular}
\caption{Performance comparison of \textbf{Micro-model-light} and \textbf{Micro-model-dense} on a test set of 1,149 utterances of 3,646.3 seconds total duration.}
\label{tab:model_comparison}
\end{table}



\section{Results and discussion}

Table \ref{tab:model_comparison} presents the performance comparison between two micro-models. The models are tested on a dataset comprising of 1,149 utterances with a cumulative duration of 3,646 seconds. We assess the models for CPU and GPU based inference tasks. The CPU used for testing is an AMD 16-Core Processor 2.9 GHz CPU with 263 GB RAM. The GPU used is an NVIDIA A100 with 80GB VRAM. Micro-model-light demonstrates a Word Error Rate (WER) of 2.65\% and a smaller model size, resulting in a faster decoding time and lower latency measured in terms of the Real-Time Factor (RTF\footnote{The Real-Time Factor (RTF) metric used to measure the speed of a system that processes an input signal (such as audio) in real time. Values under 1.0 are preferred.  Please see the Appendix for more information on RTF.}). By contrast, Micro-model-dense achieves a lower WER of 1.84\% but has a larger model size, resulting slower decoding, specially on the CPU. 

We also compare our models to three popular, pre-trained models: Whisper-Small, Whisper-Large, and Google-STT (Speech-to-text); Whisper is the best-of-breed of open source ASR engines while Google-STT is the best-of-breed for commercial ASR engines.  The results are shown in  Table \ref{tab:wer}. For the Google-STT, we specifically configured the context to transcribe only numeric content. This tailored configuration allowed us to assess the performance of the Google Speech API in accurately transcribing numerical sequences, providing insights into its suitability for applications involving voice-enabled transactions, numerical data processing, and similar scenarios. Whisper-Small is a 244 million parameter model that requires at least 2 GB VRAM, while Whisper-Large is a 1.5 billion parameter model requiring about 11 GB VRAM.  
As can be observed in the table, the WER of our micro models are superior to those of the commercial engine (Google-STT) and the open source alternative (Whisper).  In terms of RTF, we note that the RTF of our micro-models shown in the table are at least three orders of magnitude less than the RTF of Whisper, which we measured at 0.84 for Whisper-Small and 0.89 for Whisper-Large.  We were not able to measure the RTF for Google-STT because it is a cloud service, using a RESTful API interface, and in such a setting, the network latency will dominate the RTF.

\begin{table}[h]
\centering
\begin{tabular}{lcc}
\toprule
\multicolumn{1}{c}{\textbf{Model}} & \textbf{\begin{tabular}[c]{@{}c@{}}WER \\ (\%)\end{tabular}} & \textbf{\begin{tabular}[c]{@{}c@{}}RTF (meas-\\ ured on GPU)\end{tabular}} \\
\midrule
Whisper-Small                      & 5.8                                                          & 0.84                                                                       \\
Whisper-Large                      & 4.6                                                          & 0.89                                                                       \\
Google-STT                         & 2.9                                                          & N/A                                                                        \\
Micro-model-light                  & \textbf{2.6}                                                 & \textbf{0.00061}                                                           \\
Micro-model-dense                  & \textbf{1.8}                                                 & \textbf{0.00060}                                                          \\
\bottomrule
\end{tabular}
\caption{Word Error Rates}
\label{tab:wer}
\end{table}

Both of our micro models outperform the commercial engines. This indicates that small vocabulary micro models have comparable, or better accuracy compared to large vocabulary domain-general models. It is noteworthy that commercial engines are trained on hundreds of thousands of hours of audio and require sizable  resources to perform decoding. Whisper is a good example, it is trained on 680,000 hours of audio \cite{radford2023robust} over a period of many days on a cluster of NVIDIA A100 GPUs.  The large version of Whisper occupies nearly 11 GB VRAM on a GPU compared to the 0.66 GB of VRAM for our micro-models. By contrast, our micro models require low memory resources to run, many orders of magnitude less training data and take less training time.  (Our model training time was about 1 hour on 1xA100 NVIDIA GPU.)  An added advantage of a low-memory footprint is multiple models can be loaded simultaneously in memory and an application can choose the appropriate model to use for decoding.  For example, a credit card CVV consists of three numbers requiring less variation in pronouncing the three numbers.  Most users when prompted to say the CVV will utter each number independently and in sequence.  Therefore, a 3-digit CVV model that is highly trained to recognize individual numbers spoken in sequence can be used to collect the CVV while a different model can be used to collect the ZIP code where there is more variability in uttering the five numbers corresponding to the ZIP code.

To better understand the difference in WER between each model, we conducted a word-error analysis summarized in Table \ref{tab:word_errors}. We found that Google, at times, fails to capture single-digit utterances and decodes the short utterances as blank outputs. However, we found that the Google model performs well on multidigit utterances.  Whisper-Small and Whisper-Large also exhibited frequent errors on single- and two-digit utterances, with these errors becoming less frequent for multi-digit numbers.  Making more errors on short duration utterances is a known deficiency of Whisper models, since it is trained on longer audio segments (30s) making decoding shorter utterances challenging as the audio may not have enough context \cite{radford2023robust}.  Clearly, our micro-models do not suffer from this problem and are highly accurate for short audio utterances.

In closing, we note that our micro-models are not general-purpose models; they are highly specific to a particular domain, namely digit recognition.  Thus, using them in any other domain will cause high WER.

\begin{table*}[h]
\begin{tabular}{|l|l|l|l|}
\hline
\textbf{Error Type}                                                                                & \textbf{Google STT}                                                                                              & \textbf{Whisper-Small/Large}                                                                                                     & { \textbf{\begin{tabular}[c]{@{}l@{}}Micro-model\\ dense/light\end{tabular}}}                                                       \\ \hline
\begin{tabular}[c]{@{}l@{}}Single-digit (0-9) \\ transcription and \\ detection errors\end{tabular} & \begin{tabular}[c]{@{}l@{}}Frequent blank outputs \\ and occasional errors\\ (e.g. "Two" as "too" )\end{tabular} & \begin{tabular}[c]{@{}l@{}}Detection with more \\ frequent errors \\ (e.g. "three" as "spree", \\ "nine" as "fine")\end{tabular} & { \textbf{\begin{tabular}[c]{@{}l@{}}Occasional blank outputs\\  \\ \end{tabular}}}                 \\ \hline
\begin{tabular}[c]{@{}l@{}}Two-digit (11-99) \\ transcription errors\end{tabular}                   & \begin{tabular}[c]{@{}l@{}}Occasional errors\\ (e.g. "Twelve" as "well")\end{tabular}                            & \begin{tabular}[c]{@{}l@{}}More frequent errors\\ (e.g. "sixteen" as "sixty")\end{tabular}                                       & { \textbf{Highly accurate}}                                                                                                         \\ \hline
Multi-digit numbers\textgreater{}99 errors                                                         & Highly accurate                                                                                                  & \begin{tabular}[c]{@{}l@{}}Occasional errors\\ (e.g. "hundred" as "dread",\\  "thousand" as "housing")\end{tabular}             & { \textbf{\begin{tabular}[c]{@{}l@{}}Highly accurate\\ except rarely \\ missing short words \\ like "and", "two"etc.\end{tabular}}} \\ \hline
\end{tabular}
\caption{Error analysis of Automatic Speech Recognition systems: Google STT exhibits single-digit errors, Whisper-Small/Large shows increased errors for two-digit numbers, and micro-models demonstrate high accuracy with occasional multi-digit errors, particularly in recognizing short words}
\label{tab:word_errors}
\end{table*}

\section{Conclusion}
In this work, we present a domain-specific Kaldi ASR micro-model for accurate five-digit number recognition, which can outperform large, domain-general ASR models. We also present a diverse dataset for spoken digit recognition which adds to the existing resources in this domain, especially for high-privacy use cases like financial transactions where credit card or account numbers cannot be used directly for training. We trained micro-models that give low WER and highlighted the tradeoffs between the density of network, space occupied by the model and time for decoding the utterances.   Both of our micro-models outperformed the best-of-breed commercial and open-source ASR systems while using less compute resources.  Due to their inherent agility and low compute requirements, the micro-models can be run on-premises, thereby alleviating the need for sending sensitive data to a third party for decoding and transcription.  These efforts underline the utility of micro-models in specialized and lightweight ASR applications, offering efficient and accurate solutions for specific tasks. 

Future work will involve extracting digits from other data sources and improving the model's performance on out-of-distribution audio sequences. Moreover, our approach can also be applied to noisy datasets. Recognizing the significance of channel characteristics in real-world scenarios, our methodology lays the groundwork for future investigations into incorporating training audio that closely mimics the production environment. This avenue of exploration holds the potential to further enhance the robustness of our model, making it well-suited for applications in diverse and challenging acoustic conditions.

\bibliography{cite.bib}

\begin{thebibliography}{18}
\providecommand{\natexlab}[1]{#1}

\bibitem[{aur()}]{auroradataset}
 ????
\newblock Aurora Dataset.
\newblock \url{http://aurora.hsnr.de/download.html}.

\bibitem[{mni()}]{mnist_speech}
 ????
\newblock MNIST Speech Dataset.
\newblock \url{https://www.tensorflow.org/datasets/catalog/mnist\_speech}.

\bibitem[{Bastianelli et~al.(2020)Bastianelli, Vanzo, Swietojanski, and
  Rieser}]{bastianelli2020slurp}
Bastianelli, E.; Vanzo, A.; Swietojanski, P.; and Rieser, V. 2020.
\newblock SLURP: A Spoken Language Understanding Resource Package.
\newblock In \emph{Proceedings of the 2020 Conference on Empirical Methods in
  Natural Language Processing (EMNLP)}.

\bibitem[{Bekal et~al.(2021)Bekal, Shenoy, Sunkara, Bodapati, and
  Kirchhoff}]{bekal2021remember}
Bekal, D.; Shenoy, A.; Sunkara, M.; Bodapati, S.; and Kirchhoff, K. 2021.
\newblock Remember the context! ASR slot error correction through memorization.
\newblock In \emph{2021 IEEE Automatic Speech Recognition and Understanding
  Workshop (ASRU)}, 236--243. IEEE.

\bibitem[{de~Vos and Verberne(2023)}]{de2023political}
de~Vos, H.; and Verberne, S. 2023.
\newblock Political corpus creation through automatic speech recognition on EU
  debates.
\newblock \emph{arXiv preprint arXiv:2304.08137}.

\bibitem[{Dong et~al.(2023)Dong, Ding, Zhai, and Zhou}]{dong2023speech}
Dong, Z.; Ding, Q.; Zhai, W.; and Zhou, M. 2023.
\newblock A Speech Recognition Method Based on Domain-Specific Datasets and
  Confidence Decision Networks.
\newblock \emph{Sensors}, 23(13): 6036.

\bibitem[{Fredrikson, Jha, and Ristenpart(2015)}]{10.1145/2810103.2813677}
Fredrikson, M.; Jha, S.; and Ristenpart, T. 2015.
\newblock Model Inversion Attacks That Exploit Confidence Information and Basic
  Countermeasures.
\newblock In \emph{Proceedings of the 22nd ACM SIGSAC Conference on Computer
  and Communications Security}, CCS '15, 1322–1333. New York, NY, USA:
  Association for Computing Machinery.
\newblock ISBN 9781450338325.

\bibitem[{Jha(2021)}]{jha2021learning}
Jha, S. 2021.
\newblock Learning Domain Specific Language Models for Automatic Speech
  Recognition through Machine Translation.
\newblock \emph{arXiv preprint arXiv:2110.10261}.

\bibitem[{Lugosch et~al.(2021)Lugosch, Papreja, Ravanelli, Heba, and
  Parcollet}]{lugosch2021timers}
Lugosch, L.; Papreja, P.; Ravanelli, M.; Heba, A.; and Parcollet, T. 2021.
\newblock Timers and such: A practical benchmark for spoken language
  understanding with numbers.
\newblock \emph{arXiv preprint arXiv:2104.01604}.

\bibitem[{Lugosch et~al.(2019)Lugosch, Ravanelli, Ignoto, Tomar, and
  Bengio}]{lugosch2019speech}
Lugosch, L.; Ravanelli, M.; Ignoto, P.; Tomar, V.~S.; and Bengio, Y. 2019.
\newblock Speech model pre-training for end-to-end spoken language
  understanding.
\newblock In \emph{Interspeech}.

\bibitem[{Povey et~al.(2011)Povey, Ghoshal, Boulianne, Burget, Glembek, Goel,
  Hannemann, Motlicek, Qian, Schwarz et~al.}]{povey2011kaldi}
Povey, D.; Ghoshal, A.; Boulianne, G.; Burget, L.; Glembek, O.; Goel, N.;
  Hannemann, M.; Motlicek, P.; Qian, Y.; Schwarz, P.; et~al. 2011.
\newblock The Kaldi speech recognition toolkit.
\newblock In \emph{IEEE 2011 workshop on automatic speech recognition and
  understanding}, CONF. IEEE Signal Processing Society.

\bibitem[{Radford et~al.(2023)Radford, Kim, Xu, Brockman, McLeavey, and
  Sutskever}]{radford2023robust}
Radford, A.; Kim, J.~W.; Xu, T.; Brockman, G.; McLeavey, C.; and Sutskever, I.
  2023.
\newblock Robust speech recognition via large-scale weak supervision.
\newblock In \emph{International Conference on Machine Learning}, 28492--28518.
  PMLR.

\bibitem[{Ravindra and Grama(2021)}]{10.1145/3448016.3457234}
Ravindra, V.; and Grama, A. 2021.
\newblock De-Anonymization Attacks on Neuroimaging Datasets.
\newblock In \emph{Proceedings of the 2021 International Conference on
  Management of Data}, SIGMOD '21, 2394–2398. New York, NY, USA: Association
  for Computing Machinery.
\newblock ISBN 9781450383431.

\bibitem[{Roger, Farinas, and Pinquier(2022)}]{roger2022deep}
Roger, V.; Farinas, J.; and Pinquier, J. 2022.
\newblock Deep neural networks for automatic speech processing: a survey from
  large corpora to limited data.
\newblock \emph{EURASIP Journal on Audio, Speech, and Music Processing},
  2022(1): 19.

\bibitem[{Saade et~al.(2019)Saade, Coucke, Caulier, Dureau, Ball, Bluche,
  Leroy, Doumouro, Gisselbrecht, Caltagirone et~al.}]{saade2019spoken}
Saade, A.; Coucke, A.; Caulier, A.; Dureau, J.; Ball, A.; Bluche, T.; Leroy,
  D.; Doumouro, C.; Gisselbrecht, T.; Caltagirone, F.; et~al. 2019.
\newblock Spoken language understanding on the edge.
\newblock \emph{NeurIPS Workshop on Energy Efficient Machine Learning and
  Cognitive Computing}.

\bibitem[{Xu, Cohn, and Ohrimenko(2023)}]{inbook}
Xu, Q.; Cohn, T.; and Ohrimenko, O. 2023.
\newblock \emph{Fingerprint Attack: Client De-Anonymization in Federated
  Learning}.
\newblock ISBN 9781643684369.

\bibitem[{Yin et~al.(2023)Yin, Liu, Li, Guo, and Wang}]{yin2023defeating}
Yin, H.; Liu, Y.; Li, Y.; Guo, Z.; and Wang, Y. 2023.
\newblock Defeating deep learning based de-anonymization attacks with
  adversarial example.
\newblock \emph{Journal of Network and Computer Applications}, 220: 103733.

\bibitem[{Zechner and Waibel(2000)}]{zechner2000minimizing}
Zechner, K.; and Waibel, A. 2000.
\newblock Minimizing word error rate in textual summaries of spoken language.
\newblock In \emph{1st Meeting of the North American Chapter of the Association
  for Computational Linguistics}.

\end{thebibliography}

\section*{Appendix}
\vspace{2mm}

\subsection{Word Error Rate (WER)}

The WER \cite{zechner2000minimizing} is a widely accepted standard measure of ASR performance; it is expressed as a value between [0, 1.0] or as a percentage.  ASR systems seek to minimize the WER.  It is represented as the ratio of the number of edits required to transform a hypothesis string into a reference string to the total number of words in the reference string, or 

\begin{equation}
    \textup{WER = } \frac{S+D+I}{N}
\end{equation}

\noindent where S = number of substitutions required to change the hypothesis string to the reference string, D = number of deletions required, I = number of insertions, and N = total number of words in the reference string.  Lower values of WER are preferred since they indicate an ASR model that makes less errors.

\vspace{2mm}

\subsection{Real-Time Factor (RTF)}

The Real-Time Factor (RTF) metric used to measure the speed of a system that processes an input signal (such as audio) in real time.  Since ASR systems process speech and produce a transcript, the RTF can be defined as

\begin{equation}
    \textup{RTF = }\frac{T}{D}
\end{equation}

\noindent where $T =$ Time to transcribe the audio file and $D = $ Duration of the audio file.  Values of RTF $<$ 1.0 are preferred since values $\ge 1.0$ indicate that the decoding (transcribing) an audio file takes a larger amount of time than the duration of the audio itself.

\subsection{Aurora-5}

Aurora-5, introduced in 2006, extends the Aurora dataset series, focusing on hands-free speech input in real-world settings like rooms and cars. The dataset investigates additional distortion effects, including additive background noise and the impact of transmitting speech over a cellular network. Built upon the TIDigits database, Aurora-5 offers a valuable resource for assessing speech recognition system performance in complex, everyday scenarios, contributing to the refinement of speech processing technologies in telecommunications. Its exploration of hands-free input and cellular transmission adds a practical dimension to evaluating automatic speech recognition systems.

\end{document}